\documentclass[lettersize,journal]{IEEEtran}
\usepackage{amsmath,amsfonts}
\usepackage{algorithmic}
\usepackage{array}
\usepackage[caption=false,font=normalsize,labelfont=sf,textfont=sf]{subfig}
\usepackage{textcomp}
\usepackage{stfloats}
\usepackage{url}
\usepackage{amssymb}
\usepackage{multirow}
\usepackage{booktabs} 
\usepackage{verbatim}
\usepackage{graphicx}
\def\BibTeX{{\rm B\kern-.05em{\sc i\kern-.025em b}\kern-.08em
    T\kern-.1667em\lower.7ex\hbox{E}\kern-.125emX}}
\usepackage{balance}
\begin{document}
\title{Dual-perspective Cross Contrastive Learning in Graph Transformers}
\author{
    Zelin Yao, Chuang Liu, Xueqi Ma, Mukun Chen, Jia Wu, Xiantao Cai, Bo Du, Wenbin Hu
\thanks{Corresponding author: Wenbin Hu}
\thanks{Zelin Yao, Chuang Liu, Mukun Chen, Xiantao Cai, Bo Du, Wenbin Hu are with the School of Computer Science, Wuhan University, Hubei, China (E-mail: zelinyao@whu.edu.cn; chuangliu@whu.edu.cn; cmk0910@whu.edu.cn; dubo@whu.edu.cn; hwb@whu.edu.cn).}
\thanks{Xueqi Ma is with the School of Computing and Information Systems, The University of Melbourne, Melbourne, Australia (E-mail: xueqim@student.unimelb.edu.au).}
\thanks{Jia Wu is with the School of Computing, Macquarie University, Sydney, NSW 2109, Australia (E-mail: jia.wu@mq.edu.au).}
\thanks{The work of Wenbin Hu was supported by the National Key Research and Development Program of China (2023YFC2705700). This work was supported in part by the Natural Science Foundation of China (No. 82174230), Artificial Intelligence Innovation Project of Wuhan Science and Technology Bureau (No. 2022010702040070), Natural Science Foundation of Shenzhen City (No. JCYJ20230807090211021).}
}

\markboth{Dual-perspective Cross Contrastive Learning in Graph Transformers,~Vol.~18, No.~9, May~2024}%
{}

\maketitle

\begin{abstract}
Graph contrastive learning (GCL) is a popular method for leaning graph representations by maximizing the consistency of features across augmented views. Traditional GCL methods utilize single-perspective (\textit{i.e.} data or model-perspective) augmentation to generate positive samples, restraining the diversity of positive samples. In addition, these positive samples may be unreliable due to uncontrollable augmentation strategies that potentially alter the semantic information. To address these challenges, this paper proposed a innovative framework termed dual-perspective cross graph contrastive learning (DC-GCL), which incorporates three modifications designed to enhance positive sample diversity and reliability: \textbf{1)} We propose dual-perspective augmentation strategy that provide the model with more diverse training data, enabling the model effective learning of feature consistency across different views. \textbf{2)} From the data perspective, we slightly perturb the original graphs using controllable data augmentation, effectively preserving their semantic information. \textbf{3)} From the model perspective, we enhance the encoder by utilizing more powerful graph transformers instead of graph neural networks. Based on the model's architecture, we propose three pruning-based strategies to slightly perturb the encoder, providing more reliable positive samples. These modifications collectively form the DC-GCL's foundation and provide more diverse and reliable training inputs, offering significant improvements over traditional GCL methods. Extensive experiments on various benchmarks demonstrate that DC-GCL consistently outperforms different baselines on various datasets and tasks.\footnote{The codes and detailed parameter settings are available at \url{https://github.com/Celin-Yao/DC-GCL}.}
\end{abstract}

\section{Introduction}
\IEEEPARstart{G}{raph} Contrastive Learning (GCL) emerges as a novel self-supervised learning approach to that addresses the lack of labeled data in real-world scenarios. According to the general structure in contrastive learning in computer vision domain~\cite{simclr,unsupervised}, GCL methods generate two augmentaion views for each graph, aiming to maximize the feature consistency across augmented views. As a result, GCL can effectively capture the feature consistency and learn the representations of graph data, and further be demonstrated under unsupervised, semi-supervised, and transfer learning tasks. The GCL method has been widely be applied to various scenarios such as recommendation systems~\cite{xsimgcl, neigh} and molecular structures~\cite{mole-contrast, geomgcl, mocl}.

\begin{figure*}[!t] % !htb
\begin{center}
\includegraphics[width=0.98\linewidth]{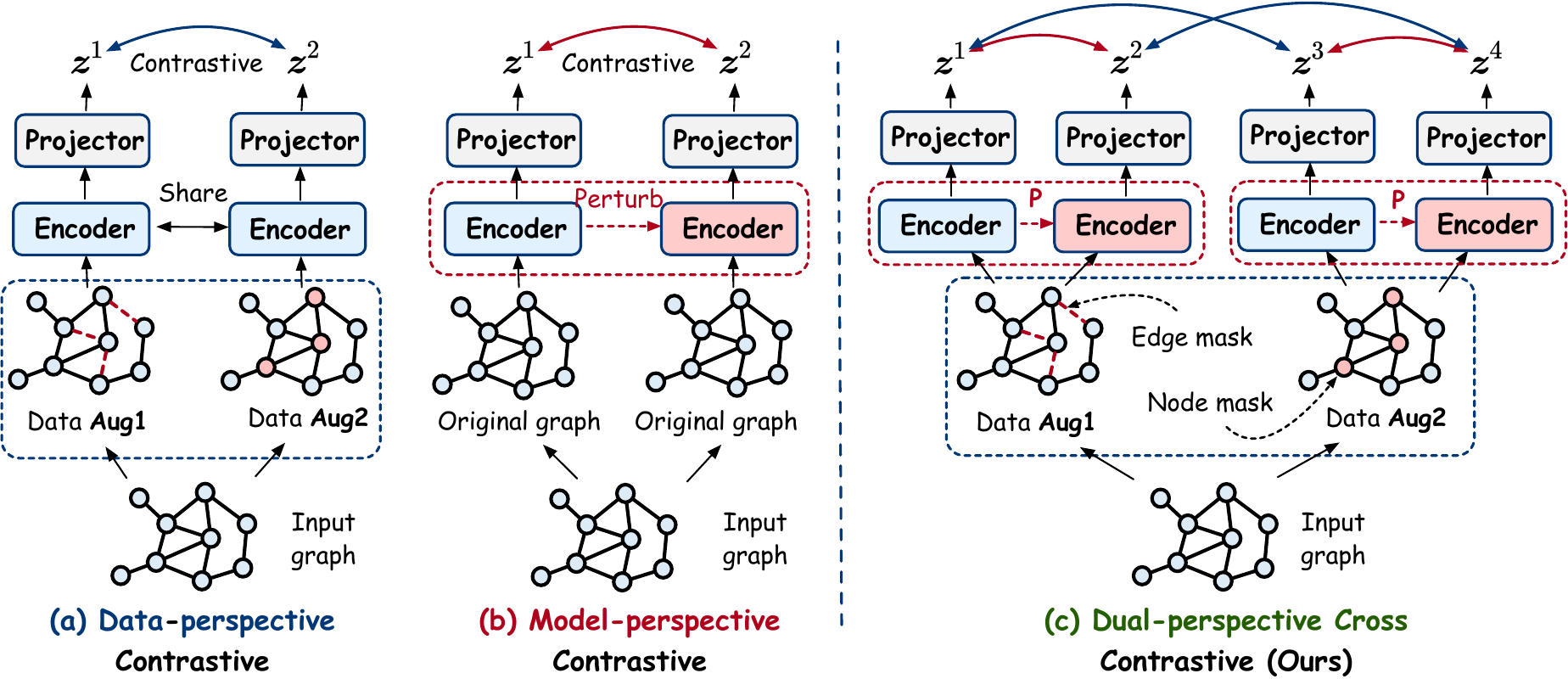}
\end{center}
\caption{\textbf{Comparison of contrastive learning methods with proposed DC-GCL.} Our DC-GCL introduces a comprehensive GT-based contrastive learning method which uses dual-perspective augmentation strategy. These novel designs significantly increase the generation of positive samples, with the aim of improving the model's ability to learn representations across diverse datasets.
}
\label{fig:method}
\end{figure*}

In popular GCL methods, the contrastive learning objects are positive samples obtained using augmentation and passed through a GNN-based encoder. Based on the augmented object, the methods can be roughly divided into two main categories: \textbf{1) Data-perspective Constrastive:} As depicted in Fig.~\ref{fig:method}(a) approaches such as GraphCL~\cite{graphCL} and G-Mixup~\cite{graphmix} modifying the original graph to create augmented graphs. Then, it places the two augmented graphs into the same encoder to obtain two correlated views. \textbf{2) Model-perspective Constrastive:} As illustrated in Fig.~\ref{fig:method}(b), methods such as SimGRACE~\cite{simgrace} operate by introducing perturbations to the encoder, which adds Gaussian noise to the model encoder. Then, it uses the original graph as the input and utilizes two encoders. One encoder is for the original data and the other for its perturbed version, resulting in two correlated views.  

During our analysis of the existing GCL framework, we identified several key insights: \textbf{1) Limited positive samples:} Current GCL methods typically perform augmentation from a single perspective, resulting in only one positive sample pair (\textit{i.e.} through data augmentation~\cite{graphCL} or model augmentation~\cite{simgrace}). However, if the generation of positive samples is confined to a single perspective, the resulting lack of diversity restricts the effectiveness of model training. \textbf{2) Unreliable augmented graphs:} In previous GCL methods, data augmentation strategies~\cite{graphCL,graphmix} focused on random modifications to the original graph. These modifications typically include the random addition or deletion of nodes and edges. This introduces diversity to the dataset, while also damaging the semantic information. Given these observations, a pertinent question arises: Is it possible to simultaneously generate diverse and reliable positive samples from two distinct augmentation perspectives, thereby enabling the model to better learn feature consistency?

To address these issues, we introduce Dual-perspective Cross Graph Contrastive Learning (DC-GCL), an innovative GT-based GCL method. We introduce three modifications in addition to traditional GCL architectures. \textbf{1) Dual-perspective Cross Contrastive:} DC-GCL incorporates a comprehensive version of data and model augmentation, surpassing the single-perspective augmentation typical in previous works. As shown in Fig.~\ref{fig:method}(c), we generate an increased number of correlated views within the DC-GCL framework. Then, the correlated views are categorized into two groups from same data or model augmentation. After that, DC-GCL computes the contrastive loss using the views in each group. It ensures that these views are closely distributed in the feature space, which guides the model to learn a more stable and reliable representation. According to the figure, views connected by the red line originate from the same data augmentation, while those connected by the blue represent the same model augmentation. \textbf{2) Controllable Data Augmentation:} Instead of relying on randomness, our approach emphasizes using a controllable strategy to obtain augmented graphs, which can effectively preserve semantic information. To achieve this, we design our data augmentation strategy by slightly perturbing the original graphs such as dropping these unnecessary nodes determined by learnable scores. Besides, we innovative proposed to use pre-trained model to generate positive samples without directly modifying the original graph structure. In summary, our data augmentation strategies are relatively controllable, ensuring that semantic information is effectively maintained. \textbf{3) Enhanced Model and Pruning-based Model Augmentation:} Unlike traditional GCL methods that predominantly use GNNs as encoders, DC-GCL employs GraphGPS~\cite{graphgps} as its core encoding mechanism. This change leverages the self-attention mechanism of transformers to enhance representation learning and to capture complex relationships within graphs effectively. Moreover, due to the GT model's redundant characteristic, we proposed pruning-based model augmentation strategies starting from the weights and architecture to obtain the positive samples. Compared to adding random noise to the encoder, pruning-based data augmentation has a slighter influence on the encoder and can generate more reliable positive samples.

To evaluate the DC-GCL method's effectiveness, we conducted unsupervised learning tasks using the TUDataset~\cite{tu-dataset} compared with 12 baseline models. Moreover, we evaluated DC-GCL's transferability by testing its performance during transfer learning tasks on MoleculeNet~\cite{moleculenet}. The experimental results consistently demonstrate that DC-GCL outperforms state-of-the-art GCL methods on the majority of datasets, showcasing remarkable improvements in both stability and accuracy. This improvement highlights the advantages of integrating augmentation with the use of GT as encoders, which are more effective in extracting representations from graphs compared to existing GCL methods. We highlight our contributions as follows:

\begin{enumerate}
  \item We propose a novel GCL method named DC-GCL which is a comprehensive model that integrates the data and model augmentation. The implementation of dual-perspective augmentation enhances the diversity of the generated positive samples, allowing the model to capture consistent graph representations across varied graph scenarios more effectively. 
  
  \item The DC-GCL method introduces refined strategies for both data and model augmentation: Controllable Data Augmentation: Unlike traditional methods that rely on randomness, our approach ensures that data augmentation is controlled, preserving the graphs' semantic integrity. This method mitigates the risk of distorting the graphs' properties, augmented samples' quality and reliability. In addition, DC-GCL utilizes GTs as encoders, and based on the architecture, we propose Pruning-Based Model Augmentation: Recognizing the redundancy often present in GTs, we develop three pruning-based strategies for model augmentation. These strategies are designed to generate more reliable positive samples by reducing unnecessary complexity during the encoding process.

  \item Finally, we conduct extensive experiments to compare DC-GCL with 12 GT-based supervised, graph kernel and contrastive-based baseline models across two graph classification tasks on various real-world datasets. The experimental results consistently validate the proposed method's effectiveness.
  
\end{enumerate}
\section{Related Work}
\textbf{Graph Contrastive Learning.} Contrastive learning primarily involves utilizing an embedding space to represent data samples. It also involves bringing similar samples closer while pushing dissimilar ones farther away without supervision signals. This method performs exceptionally well in both computer vision~\cite{simclr, momentum} and natural language process~\cite{sg-bert, cline} fields, achieving effective data representation and various downstream tasks. Inspired by contrastive learning in other domains, graph contrastive learning (GCL) is also advancing gradually. The GCL method can be roughly divided into two augmentation perspectives. Within the data augmentation domain, methods such as the GRACE~\cite{GRACE} introduces two data augmentation strategies, including removing edges and masking node features, and focuses on addressing node-level tasks. GraphCL~\cite{graphCL} asserts that adopting various data augmentation methods leads to variations in the data from different categories. JOAO~\cite{joao} enhances GraphCL by enabling it to automatically select its data augmentation strategies. SFA~\cite{sfa} leverages the spectral information of graphs to generate data transformations. Similarly, GraphAug~\cite{graphaug} obtains augmented graphs by performing a series of automated selection and learned transformation operations. NodeMixup~\cite{nodemixup} adopts the labeled-unlabeled pairs mixup strategy where labeled nodes are mixed with unlabeled ones to synthetically expand training data and improve GNN model generalization. Moreover, DRGCL~\cite{drgcl} obtains augmented graphs by randomly preserving specific graph representation  dimensions and adjusting them using learnable dimensional weights. In comparison, methods within the model augmentation domain such as SimGRACE~\cite{simgrace} augment the model by introducing Gaussian noise, thereby avoiding manual trial and error. Ma-GCL~\cite{magcl} manipulates the neural architectures of GNN view encoders using asymmetric, random, and shuffling strategies instead of perturbing graph inputs or model parameters. In contrast to these models, DC-GCL proposes the dual-perspective augmentation strategy, which integrates both data and model augmentation, aiming to the surpass the state-of-the-art GCL framework by increasing positive samples.

\textbf{Graph Transformers.} Although it was initially designed for natural language processing tasks, the transformer's~\cite{transformer} architecture has shown remarkable flexibility and effectiveness across various domains. GROVER~\cite{grover} first proposed the graph transformers (GTs) model, which incorporates message passing networks into a transformer-style architecture, used for acquiring extensive unlabeled molecular information. Thereafter, GTs gradually became popular in the application of graph-structured data. As a result, various approaches utilize GT-based models to tackle graph-level tasks, as they exhibit significant improvements compared to GNNs. GT~\cite{graph-transformer} utilizes Laplacian eigenvectors as positional encoding (PE), leveraging their PE to obtain richer information compared to GNNs. Graphormer~\cite{graphormer-v1} modifies attention scores using structural information. Many existing GNNs can be considered as special cases of Graphormer, showcasing its outstanding performance across various datasets. GraphTrans~\cite{graphtrans} strives to combine both local and global information, offering greater flexibility compared to GNNs. GraphGPS~\cite{graphgps} provides a GT framework that combines local and global information, integrating them to produce the model's output. GRIT~\cite{grit} incorporates effective graph inductive bias without using GNN-based message passing neural networks. GPTrans~\cite{gpt} proposed a novel attention mechanism, passing information between nodes and edges in three ways: node-to-node, node-to-edge, and edge-to-node. In addition, Gapformer~\cite{gapformer} enhances node classification by combining GTs with pooling to reduce complexity and noise. Finally, Gradformer~\cite{gradformer} employed a decay mask on attention heads to enhance the model's focus on local and global information. Generally, these studies were all performed using end-to-end supervised learning, which requires a substantial amount of labeled data and time. Incorporating GT into contrastive learning can mitigate this issue. This can be achieved by pre-training the model using unlabeled data, which can be transferred to the downstream tasks easily.

\section{Preliminaries}
\textbf{Notation.} A graph $\mathcal{G}(\boldsymbol{A}; \boldsymbol{X})$ consists an adjacency matrix $\boldsymbol{A} \in \{0, 1\}^{ n \times n}$ and a node feature matrix $\boldsymbol{X} \in \mathbb{R}^{ n \times d}$, where $n$ denotes the number of nodes, $d$ represents the node feature dimension, and $\boldsymbol{A}[i, j]=1$ indicates the presence of an edge between nodes $v_{i}$ and $v_{j}$, otherwise $\boldsymbol{A}[i, j]=0$.

\textbf{Graph Transformer.} GTs~\cite{transformer,graphormer-v1} comprises two essential parts: a multi-head self-attention (MHA) module and a feed-forward network (FFN). Given the node embedding matrix $\boldsymbol{H}^{(l)} \in \mathbb{R}^{ n \times d^{(l)}}$ in a graph, a single attention head is computed as follows: 
\begin{equation}
\boldsymbol{H}^{(l+1)}=\operatorname{softmax}\left(\frac{\boldsymbol{Q}^{(l)} \boldsymbol{K}^{(l)\top}}{\sqrt{d^{(l)}}}\right) \boldsymbol{V}^{(l)}, 
\label{eq:sha}
\end{equation}
where $\boldsymbol{H}^{(l+1)} \in \mathbb{R}^{n \times d^{(l+1)}}$ is the output matrix,  $d^{(l+1)}$ is the $l+1$-layer hidden dimension, and $\boldsymbol{Q}^{(l)} \in \mathbb{R}^{n \times d^{(l)}}$, $\boldsymbol{K}^{(l)} \in \mathbb{R}^{n \times d^{(l)}}$, and $\boldsymbol{V}^{(l)} \in \mathbb{R}^{n \times d^{(l)}}$ are the query, key, and value vectors, respectively. These vectors are the projection results of $\boldsymbol{H}^{(l)} \in \mathbb{R}^{ n \times d^{(l)}}$ and are expressed as follows:
\begin{equation}
\boldsymbol{Q}^{(l)}=\boldsymbol{H}^{(l)} \boldsymbol{W}^Q; 
\boldsymbol{K}^{(l)}=\boldsymbol{H}^{(l)} \boldsymbol{W}^K; 
\boldsymbol{V}^{(l)}=\boldsymbol{H}^{(l)} \boldsymbol{W}^V ,
\end{equation}
where $\boldsymbol{W}^Q \in \mathbb{R}^{d^{(l)} \times d^{(l+1)}}, \boldsymbol{W}^K \in \mathbb{R}^{d^{(l)} \times d^{(l+1)}}$, and $\boldsymbol{W}^V \in \mathbb{R}^{d^{(l)} \times d^{(l+1)}}$ are projection matrices. Note that the above single-head self-attention module can be generalized into a MHA via the concatenation operation. 

\textbf{The Graph Contrastive Learning Architecture.} As the architecture of graph contrastive learning (GCL) first proposed by GraphCL~\cite{graphCL}, this framework can be summarized as follows: First, GCL generates positive samples through augmentation. The GCL architecture can be divided into data and model augmentation according to the positive sample generation method. Data augmentation generates the correlated graph and put it into a single encoder, while model augmentation perturbs the encoder and inserts the original graph into different encoders. Then, the encoder generates two correlated views, which serve as the positive samples. After that, as proposed by the SimCLR~\cite{simclr}, $g(\cdot)$ is adopted as a non-linear projection which can map representations $\boldsymbol{h}$ and $\boldsymbol{h'}$ into another latent space before calculating the contrastive loss. Typically, a two-layer perceptron will be utilized to obtain the final views $\boldsymbol{z}$ and $\boldsymbol{z'}$,
\begin{equation}
    \boldsymbol{z} = g(\boldsymbol{h}), \quad \boldsymbol{z'} = g(\boldsymbol{h'}).
\end{equation}
Finally, normalized temperature-scaled cross entropy loss (NT-Xent) is applied as the loss function to maximize feature consistency across the positive pairs $\boldsymbol{z}$ and $\boldsymbol{z'}$. This is performed according to previous studies~\cite{nt-xent,graphCL,simgrace}. Additional, the final loss is obtained by adding the NT-Xent of all the positive sample pairs.
 
\section{Method}
\label{sec:method}
\subsection{Overview}
In this section, we delve into a comprehensive DC-GCL architecture overview, highlighting the key modifications: dual-perspective augmentation(\textbf{$\S$\ref{sec:cross}}) and multi-view contrastive loss (\textbf{$\S$\ref{sec:loss}}) to address the lack of positive sample diversity within traditional GCL methods. To further address the relative unreliability of the generated samples, we introduce controllable data (\textbf{$\S$\ref{sec:data_aug}}) and pruning-based model augmentation strategies (\textbf{$\S$\ref{sec:model_aug}}), while remaining compatible with existing augmentation strategies. Specifically, our encoder is GT-based and includes novel components such as positional encoding and attention heads. In the DC-GCL model, these components generate more reliable positive samples, enabling the model to effectively capture the consistency across different views. Finally, we provide a evaluation of the positive samples' reliability based on properties alignment and uniformity (\textbf{$\S$\ref{sec:discussion}}).
\subsection{Dual-Perspective Augmentation}
\label{sec:cross}
Contrastive learning primarily involves creating the positive samples through augmentation, aiming to move similar samples closer while pushing dissimilar ones farther apart. Existing GCL methods focus on single-perspective augmentation, resulting in two positive samples and calculating their contrastive loss. However, single-perspective augmentation restricts positive sample diversity, thereby limiting the model's ability to effectively capture the feature consistency within correlated views. Consequently, we introduce a comprehensive approach to data and model augmentation for contrastive learning (\textit{i.e.} dual-perspective augmentation). This innovative method is capable of generating twice as many positive samples as previous approaches. In the following sections~\ref{sec:data_aug} and~\ref{sec:model_aug}, we will provide a detailed overview of its implementation.  
\begin{figure*}[!h] % !htb
\begin{center}
\includegraphics[width=0.98\linewidth]{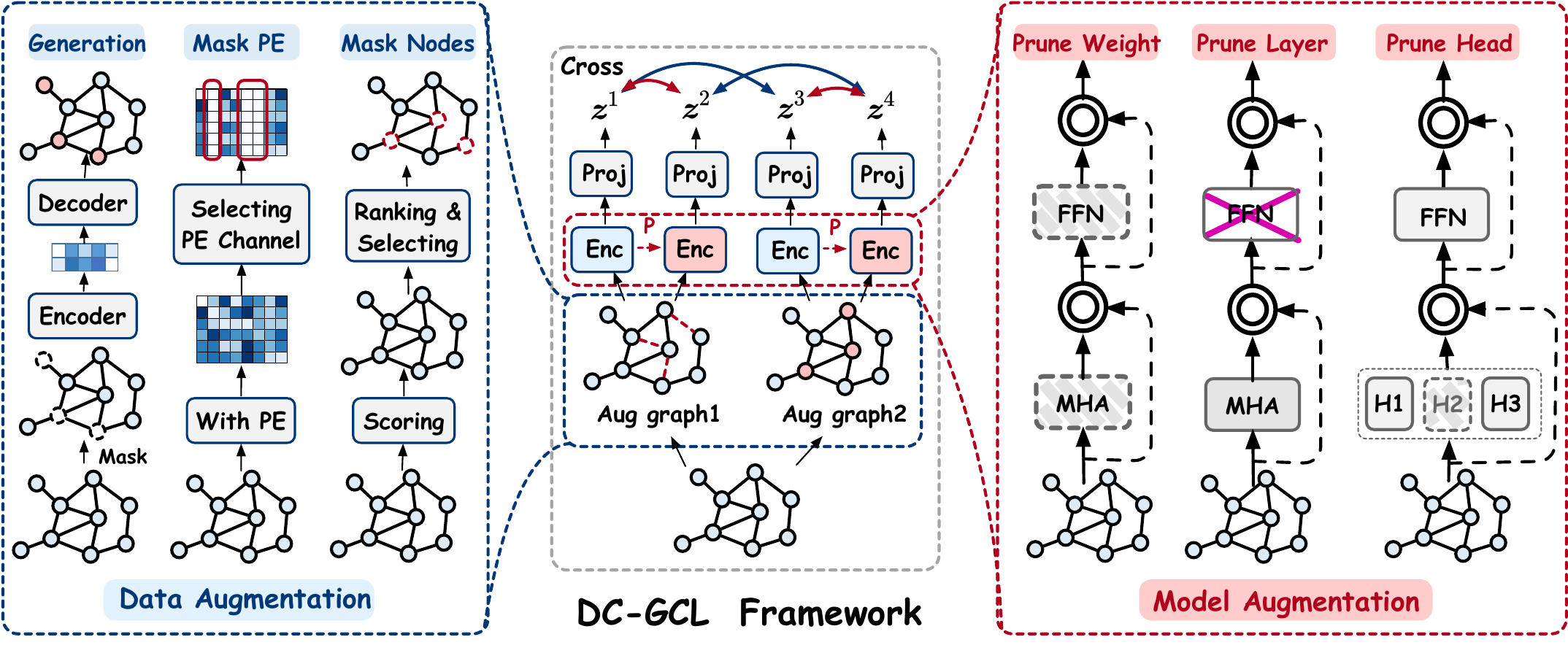}
\end{center}
\caption{\textbf{Overview the architecture of DC-GCL.} We adopt data augmentations to obtain two correlated graphs, and send these graphs to the GT-based encoder and its perturbed version to get the positive samples. During this process, we propose three controllable data augmentation and prune-based model augmentation methods respectively. A comprehensive introduction for these methods is provided in Section~\ref{sec:data_aug} and~\ref{sec:model_aug}.
}
\label{fig:model}
\end{figure*}

As depicted in the Fig.~\ref{fig:model}, we first applied the data augmentation function $\tau_d(\cdot)$ to the original graph $\mathcal{G}(\boldsymbol{A}; \boldsymbol{X})$, resulting in two correlated graphs $\mathcal{G}(\boldsymbol{A}; \boldsymbol{X})$ and $\hat{\mathcal{G}}(\hat{\boldsymbol{A}}; \hat{\boldsymbol{X}})$. Then, we applied the model augmentation function $\tau_m(\cdot)$ to the encoder, generating the origin encoder $f(\mathcal{G}; \boldsymbol{\theta})$ and the perturbed version $\hat{f}(\mathcal{G}; \hat{\boldsymbol{\theta}})$. Next, we placed the two augmented graphs into two encoders respectively, yielding four representations $\boldsymbol{h}^1, \boldsymbol{h}^2, \boldsymbol{h}^3, \boldsymbol{h}^4$. By generating two times more positive samples than previous augmentation methods, DC-GCL enhances the model's ability to capture the consistency across different views and improves its performance significantly.

\subsection{Multi-View Contrastive Loss}
\label{sec:loss}
After the dual-perspective augmentation operation, we obtained four representations $\boldsymbol{h} = \{\boldsymbol{h}^1, \boldsymbol{h}^2, \boldsymbol{h}^3, \boldsymbol{h}^4\}$. Then we adopted a two-layer perception $g(\cdot)$ as the projection layers to generate the correlated view $z$. We denoted $\boldsymbol{z}_n^1, \boldsymbol{z}_n^2, \boldsymbol{z}_n^3, \boldsymbol{z}_n^4$ as the four views for the $n$-th graph. Next, we classified the representation origins into two categories: those generated through identical data augmentation and those derived from identical model augmentation. This categorization results in four distinct groups: \textbf{1) Same data augmentation groups:} ($\boldsymbol{z}_n^1, \boldsymbol{z}_n^2$) and ($\boldsymbol{z}_n^3, \boldsymbol{z}_n^4$). Each positive sample pair in the groups originates from the same data augmentation operation, but is processed using distinct encoders $f(\cdot)$ and $\hat{f}(\cdot)$. \textbf{2) Same model augmentation groups:} ($\boldsymbol{z}_n^1, \boldsymbol{z}_n^3$) and ($\boldsymbol{z}_n^2, \boldsymbol{z}_n^4$). These sample pairs are generated from different data augmentation operations $\tau_d(\cdot)$ and $\hat{\tau}_d(\cdot)$, but have been processed with the same encoder to obtain the final representations. 

Following the architecture of previous GCL methods, we computed the NT-Xent for each positive pair group, while negative pairs were selected from the other $N - 1$ representations within the minibatch. For convenience, we renamed the two representations in the same group for the n-th graph as $\boldsymbol{z}_n$  $\boldsymbol{z}_n'$. We also denoted the cosine similarity function as sim($\boldsymbol{z}_n, \boldsymbol{z}_n'$) = ${\boldsymbol{z}_n}^{\top} {\boldsymbol{z}_n}'$/$||\boldsymbol{z}_n||||{\boldsymbol{z}_n}'||$. Finally, the loss function in a single group is defined as:
\begin{equation}
    \mathcal{L}_n = -\mathrm{log} \frac{\mathrm{exp}(\mathrm{sim}(\boldsymbol{z}_n, {\boldsymbol{z}_n}')/t)}
    {\sum_{n'=1, m!=n}^N \mathrm{exp}(\mathrm{sim}(\boldsymbol{z}_n, \boldsymbol{z}_m)/t)},
\end{equation}
where $t$ denotes the temperature parameter. The final loss was computed for all the postive pairs within the minibatch.

\subsection{Controllable Data Augmentation}
\label{sec:data_aug}
Data augmentation plays a significant role in GCL and enhances the diversity and quantity of data by applying transformations to the original data. It aims to increase the dataset diversity and improve the model's performance. The transformation is defined as follows:
\begin{equation}
\begin{aligned}
            &\tau_d(\cdot): \mathcal{G}(\boldsymbol{A}; \boldsymbol{X}) \rightarrow \hat{\mathcal{G}}(\hat{\boldsymbol{A}}; \hat{\boldsymbol{X}}), \\
    &\boldsymbol{h} = f(\mathcal{G}; \boldsymbol{\theta}), \quad \boldsymbol{h'} = f(\hat{\mathcal{G}}; \boldsymbol{\theta}),
\end{aligned}
\end{equation}
where $\tau_d(\cdot)$ denotes the data augmentation function, $f(\cdot; \boldsymbol{\theta})$ represents the GT-based encoder, and $\boldsymbol{\theta}$ denotes the model's weight. Data augmentation's main principle involves applying slight perturbations to the graph $\mathcal{G}(\boldsymbol{A}; \boldsymbol{X})$ and altering the adjacency matrix $A$ and the feature matrix $X$, resulting in a new graph $\hat{\mathcal{G}}(\hat{\boldsymbol{A}}; \hat{\boldsymbol{X}})$. However, common approaches such as randomly adding or dropping nodes and edges, can potentially damage the graph structure, leading to meaningless augmented views~\cite{free-aug}. To address this issue, we designed three controllable data augmentation strategies from two different perspectives. First, we modified the original graph structures (\textit{i.e.} \textbf{1) mask positional encoding channels} and \textbf{2) selective node masking}). Second, we proposed \textbf{3) generative-based data augmentation}, which uses a pre-trained model to generate the input graph's features without modifying the original graph. We will introduce the detailed data augmentation strategies below and depict its procedure in Fig.~\ref{fig:model}.

%  Within this framework, the incorporation of Positional Encoding (PE) is indispensable. 
\textbf{1) Positional Encoding Channels Masking.} DC-GCL adopts GTs as the model's encoder, utilizing self-attention mechanisms to capture global dependencies across the graph's nodes. In this framework, positional encoding (PE) integrates the nodes' positional information within the graph into the GTs, thereby enhancing the model's ability to capture the relative positional relationships between nodes. An intuitive approach is to apply masking operation to PE to generate augmented graphs and train the model to reconstruct these masked PEs. Compared to existing methods that randomly add or drop nodes and edges, our method does not disrupt the graph structure directly, thereby can effectively preserve the semantic information. Commonly, GTs' input features include the concatenation of PE and node features as follows:
\begin{equation}
    \tilde{\boldsymbol{X}} = \mathrm{CONCAT}(\boldsymbol{X}; \boldsymbol{H}_{pe}),
\end{equation}
where $\mathrm{CONCAT(\cdot)}$ refers to the operation of concatenation, and $\tilde{\boldsymbol{X}}$ denotes the input features. We generated the augmented graphs by applying a masking operation to PE with a certain ratio. Then, we randomly selected the PE dimensions and assigned the zero value for all nodes. The masking procedure is as follows:
\begin{equation}
\begin{aligned}
          & b_m \sim \mathrm{Bernoulli}(1 - p); \\
      & \boldsymbol{H}_{pe_i} = b_m * \boldsymbol{H}_{pe_i},
\end{aligned}
\end{equation}
where $\boldsymbol{H}_{pe_i}$ denotes PE's $i$-th dimension feature, $b_m$ represents a variable following a Bernoulli distribution and $p$ indicates the mask ratio of $\boldsymbol{H}_{pe}$. 

\textbf{2) Selective Node Masking.} Masking nodes is a common data augmentation strategy, which randomly discards the nodes and their connections. However, this strategy overlooks the importance of key nodes, which may play crucial roles in the graph. This damages the graph's semantic information of the graph (\textit{e.g.} central nodes, highly connected nodes). To address this issue, we employed a learnable approach which assesses the node significance during training process. Specifically, we introduced a masking matrix $\boldsymbol{M} \in \{0, 1\}^{n}$ to identify the masked nodes and incorporate the Gumbel distribution $\boldsymbol{D}$~\cite{gumble} into the nodes' feature $\boldsymbol{x}$, 
\begin{equation}
    \boldsymbol{D}_i = -\mathrm{log}(\mathrm{log}(\epsilon_i)), \quad \epsilon_i \sim U(0, 1),
\end{equation}
\begin{equation}
\boldsymbol{\tilde{x}}_i = \boldsymbol{x}_i + \boldsymbol{D}_i.
\end{equation}
Then, we calculated the importance score $\boldsymbol{S}$ for all the nodes, a higher score $\boldsymbol{S}_i$ indicates greater importance for the node $i$. In addition, $t$ denotes the temperature parameter. The importance score formula is expressed as follows:
\begin{equation}
    \boldsymbol{S}_i = \frac{e^{\boldsymbol{\tilde{x}}_i/t}}{\sum_{j = 1}^n{e^{\boldsymbol{\tilde{x}}_j/t}}}.
\end{equation}
The scores $\boldsymbol{S}$ are sorted in descending order, and if the $\boldsymbol{S}_i$ of node $i$ is smaller than the largest $k$ scores among all the nodes, we set $\boldsymbol{M}_i = 0$, which indicates that node $i$ is masked. Therefore, this mechanism has the capability to filter out unnecessary nodes in the graph during the training process. 

\textbf{3) Generative-Based Data Augmentation.} Traditional data augmentation strategies typically involve applying transformations directly to original graphs, risking altering the graph semantic information. In contrast, we introduce an innovative methodology for generating graph's feature matrix utilizing a pre-trained model. Our initial step is pre-training an model implemented by GraphMAE~\cite{graphmae}. Specifically, GraphMAE first randomly samples a subset of nodes $V_{[M]} \in V$ and replaces them with a learnable vector $\boldsymbol{x}_{[M]} \in \mathbb{R}^d$, we use $\mathrm{MASK(\cdot)}$ to replace this operation. Then, we inserted the graph into encoder $f_E(\cdot)$ into obtain the representation $\boldsymbol{H}_E$, 
\begin{equation}
    \boldsymbol{X}_{M} = \mathrm{MASK}(\boldsymbol{X}), \quad
    \boldsymbol{H}_E = f_E(\boldsymbol{A}; \boldsymbol{X_{M}}).
\end{equation}
Finally, we remasked the nodes within the set $V_{[M]}$ using another learnable vector $\boldsymbol{x}_{[R]} \in \mathbb{R}^{d_E}$, where $d_E$ denotes the hidden layer size. This operation is represented by $\mathrm{REMASK}(\cdot)$ represents this operation. The remasked vector is then put into the decoder function $f_D(\cdot)$ to reconstruct the masked nodes, 
\begin{equation}
    \boldsymbol{H}_R = \mathrm{REMASK}(\boldsymbol{H}_E), \quad
    \boldsymbol{\tilde{X}} = f_D(\boldsymbol{A}; \boldsymbol{H}_R).
\end{equation}
After pre-training, we sampled a subset of nodes from the original graph $\tilde{V} \subset V$ and mask them. Subsequently, we utilized the pre-trained encoder and decoder to generate a new graph $\tilde{\mathcal{G}}(\boldsymbol{A}; \tilde{\boldsymbol{X}})$. Finally, the features derived from generated graph are employed to replace those within the subset $\tilde{V}$. Thus, the augmented graph features can be represented as:
\begin{equation}
    \boldsymbol{\hat{X}}_i = 
    \begin{cases}
        \boldsymbol{\tilde{x}}_i, \quad & v_i \in \tilde{V} \\
        \boldsymbol{x}_i, \quad & v_i \notin \tilde{V} .
    \end{cases}
\end{equation}

\subsection{Pruning-based Model Augmentation}
\label{sec:model_aug}

Model-perspective augmentation involves making a perturbation on the model's encoder. We adopted the function $\tau_m(\cdot)$ to generate the perturbed encoder. Specifically, this function uses the original graph as input and employs the original and perturbed versions as the encoders to generate two correlated views $\boldsymbol{h}$ and $\boldsymbol{h'}$.
\begin{equation}
\begin{aligned}
        &\tau_m(\cdot): f(\mathcal{G}; \boldsymbol{\theta}) \rightarrow \hat{f}(\mathcal{G}; \hat{\boldsymbol{\theta}}); \\
    &\boldsymbol{h} = f(\mathcal{G}; \boldsymbol{\theta}), \quad \boldsymbol{h'} = \hat{f}(\mathcal{G}; \hat{\boldsymbol{\theta}}),
\end{aligned}
\end{equation}
where $f(\cdot; \boldsymbol{\theta})$ denotes the original encoder and $\hat{f}(\cdot; \boldsymbol{\hat{\theta}})$ is the perturbed encoder. $\boldsymbol{\theta}$ and $\boldsymbol{\hat{\theta}}$ are the weight and perturbed encoder version respectively. 

Due to the GT architecture's complexity and redundancy, we proposed pruning-based model augmentation to controllable and slightly perturb the encoder. Specifically, we have devised three distinct model augmentation methods, which are also roughly divided into two categories: modification of model's weight (\textit{i.e.} \textbf{1) prune weights}) and modification of model's architecture (\textit{i.e.} \textbf{2) prune layers} and \textbf{3) prune attention heads}). These augmentation strategies will are detailed below and illustrated in Fig.~\ref{fig:model}, highlighting these methods could reduce the complexity of the model by eliminating redundant components and parameters and produce more reliable positive samples.

\textbf{1) Weight Pruning-Based Augmentation.} Previous methods such as SimGRACE~\cite{simgrace} perturbed the encoder by adding Gaussian noise to the weight. However, this method is relatively challenging to control because randomly modifying the weight can result in critical information loss. To address this issue, we employed the L1\_unstructured pruning method, which focuses on pruning weights with the smallest L1-norm value of the model weight during training. The pruning weight process can be summarized by the following equations:
\begin{equation}
\begin{aligned}
    & \mathrm{idx} = \mathrm{TopK}(-\| \boldsymbol{\theta} \|_1, \lceil p \cdot |\boldsymbol{\theta}| \rceil); \\
    & \boldsymbol{M_{\theta}} = \mathrm{Zero}(\boldsymbol{M_{\theta}}, \mathrm{idx}), \quad \hat{\boldsymbol{\theta}} = \boldsymbol{\theta} \odot \boldsymbol{M_{\theta}},
    \end{aligned}
\end{equation}
where $\mathrm{TopK}$ returns the index of the largest $\lceil p \cdot |\boldsymbol{\theta}| \rceil$ value of $\| \boldsymbol{\theta} \|_1$, $\mathrm{Zero}(\cdot)$ is the function that sets the values of $\boldsymbol{M_{\theta}}$ with index $\mathrm{idx}$ to 0, where $\boldsymbol{M_{\theta}}=1^{|\boldsymbol{\theta}|}$ initially. $\| \boldsymbol{\theta} \|_1$ denotes the L1-norm of the model weight $\boldsymbol{\theta}$, $|\boldsymbol{\theta}|$ indicates the model weight number, $p$ is the pruning probability, and symbol $\odot$ represents the element-wise vector multiplication. Selectively pruning the weights with the smallest L1-norms eliminates the most insignificant parameters while retaining information for enhanced model performance. This process produces the perturbed encoder with the adjusted weight $\boldsymbol{\hat{\theta}}$. 
%We select the $\tilde{\theta} \subset \theta$ whose L1 norm results relatively low. By selectively pruning weights with the smallest L1 norm, we ensure that the most insignificant parameters are eliminated, while retaining the crucial information necessary for model performance. This process results in the perturbed encoder with the adjusted weight $\boldsymbol{\hat{\theta}}$. 

\textbf{2) Layers Pruning-Based Augmentation.} With GT's introduction of various complex components and parameters, the drop layer prevents over-fitting and is an effective positive sample pair generation strategy. It is possible to enable the model to learn multiple effective layers during training, instead of overly relying on the output from a single layer. Following the stochastic depth~\cite{drop-path}, we implemented the pruning layer strategy using Bernoulli random variables $b^{(l)} \in \{0, 1\}^{N_l}$ ($N_l$ is the number of layers) and residual connections, as expressed as below:
\begin{equation}
    \boldsymbol{H}^{(l+1, i)} = \mathrm{ReLU}(b^{(l)}f^{(l)}(\boldsymbol{H}^{(l, i)}) + \mathrm{Id}(\boldsymbol{H}^{(l, i)})),
\end{equation}
where $\boldsymbol{H}^{(l, i)}$ denotes the $l$-th layer, $i$-th head representations, $\mathrm{Id}(\cdot)$ is the identity transformation, and $f^{(l)}(\cdot)$ indicates the GT-based encoder within $l$-th layer, which consists of multiple multi-head attention and normalization layers. Moreover, the positive sample is generated by selecting different neural network layers during each epoch, enabling the model to effectively capture the complex structures and features within the input data.

\textbf{3) Attention Heads Pruning-Based Augmentation.} The multi-head mechanism (MHA) is essential in transformers for obtaining graph representations. Each independent attention head can capture information from various graph perspectives. Michel et al.~\cite{prunehead} stated that many of the heads in MHA mechanisms extract information that does not significantly impact the final outcome, leading to redundancy. Thus, we propose an augmentation strategy that randomly discards the attention heads' outputs with a specific probability during training. This approach slightly perturbs the representations. To determine whether this attention head should be pruned or retained, we apply the Bernoulli random variables $b_h^{(l, i)} \in \{0, 1\}^{n \times \frac{d^{(l)}}{N_h}}$ ($N_h$ is number of attention heads) for $i$-th attention head, as described below:
\begin{equation}
    \boldsymbol{H}^{(l)} = \mathrm{CONCAT}(b_h^{(l, 1)}\boldsymbol{H}^{(l, 1)} , ..., b_h^{(l, n)}\boldsymbol{H}^{(l, n)})\boldsymbol{W}^O,
\end{equation}
where $\boldsymbol{H}^{(l)}$ denotes the $l$-th layer's output representation, $\boldsymbol{H}^{(l, i)}$ indicates the $i$-th head's representation, and $\boldsymbol{W}^O \in \mathbb{R}^{d^{(l)} \times d^{(l)}}$ is the linear layer.

\subsection{Discussion} 
\label{sec:discussion}
\textbf{Alignment and Uniformity Analysis.} To evaluate the reliability of the DC-GCL-generated positive samples, we utilized the analysis tool from the study by ~\cite{align}. This tool contains two key contrastive learning properties: alignment and uniformity. These properties evaluate the quality of representations obtained from the model. Alignment is the expected distance between positive pairs. It evaluates the proximity of the positive pairs within the embedding space. Meanwhile, uniformity indicates the distribution of feature vectors within the embedding space. The alignment and uniformity properties are described as follows:
\begin{equation}
    \mathcal{L}_{\text{ali}} \triangleq \underset{{(\xi, \eta) \sim P_{\mathrm{pos}}}}{\mathbb{E}}  [{\|{f(\xi) - f(\eta)}\|_2^\alpha}], \quad \alpha > 0,
\end{equation}
\begin{equation}
    \mathcal{L}_{\text{uni}} \triangleq \mathrm{log} \underset{(\xi, \eta) |\overset{i.i.d.}{\sim} P_{\mathrm{pos}}}{\mathbb{E}} [e^{-\beta\|f(\xi)-f(\eta) \|_2^\beta }], \quad \beta > 0,
\end{equation}
where $\mathcal{L}_{\text{ali}}$ and $\mathcal{L}_{\text{uni}}$ denote the alignment and uniformity properties, respectively. $P_{\mathrm{pos}}$ indicates the distribution of positive pairs, $i.i.d$ refers to the independent and identically distributed pairs, and $\xi$ and $\eta$ are the augmented graphs originating from the same sample. 

Furthermore, we provide a comprehensive encoder version for the positive pairs generated by the data and model augmentation strategies, as described below:
\begin{equation}
    \mathcal{L}_{\text{ali}} \triangleq \underset{{(\mathcal{G}, \hat{\mathcal{G}}) \sim P_{\mathrm{pos'}}}}{\mathbb{E}}  [{\|{f(\mathcal{G}; \boldsymbol{\theta}) - \hat{f}(\hat{\mathcal{G}}; \hat{\boldsymbol{\theta}})}\|_2^\alpha}], \quad \alpha > 0.
\end{equation}
\begin{equation}
    \mathcal{L}_{\text{uni}} \triangleq \mathrm{log} \underset{(\mathcal{G}, \hat{\mathcal{G}}) |\overset{i.i.d.}{\sim} P_{\mathrm{pos'}}}{\mathbb{E}} [e^{-\beta\|f(\mathcal{G}; \boldsymbol{\theta})-\hat{f}(\hat{\mathcal{G}}; \hat{\boldsymbol{\theta}}) \|_2^\beta }], \quad \beta > 0
\end{equation}
where $P_{\mathrm{pos'}}$ is the graphs $\mathcal{G}$ and $\hat{\mathcal{G}}$ distributions. $f(\cdot; \boldsymbol{\theta})$ and $\hat{f}(\cdot; \hat{\boldsymbol{\theta}})$ denote the encoder and its perturbed version. We calculated the alignment and uniformity within the abovementioned group. First, we set the parameters as $\alpha=2$ and $\beta=2$ during the experiment. 
\begin{figure}[!t] % !htb
\begin{center}
\includegraphics[width=1.0\linewidth]{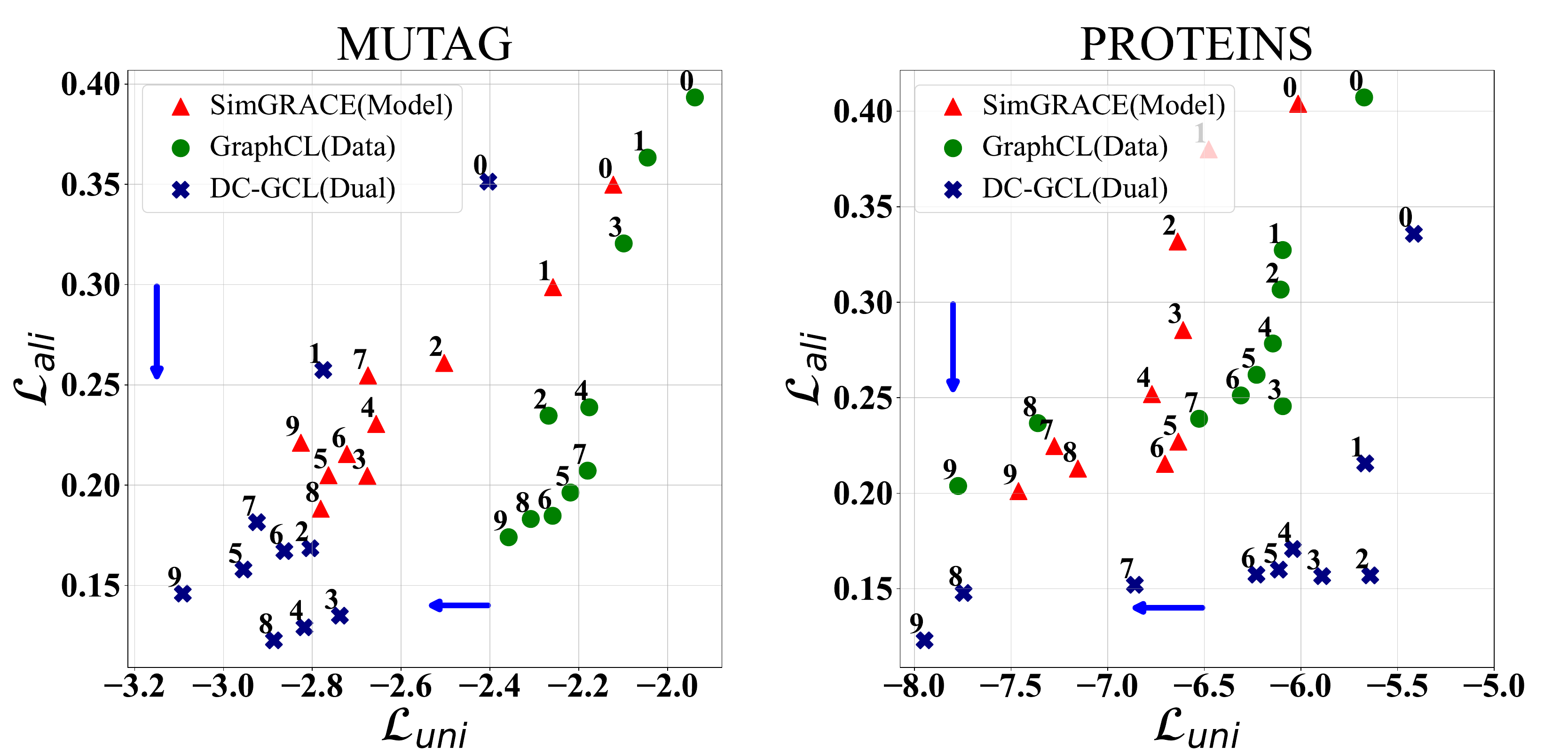}
\end{center}
\caption{$\mathcal{L}_{\text{ali}}$-$\mathcal{L}_{\text{uni}}$ visualization for GraphCL, SimGRACE, and DC-GCL for MUTAG and PROTEINS datasets. Numbers around the points in the figure represent the epochs. For both properties alignment and uniformity, the lower the better.}
\label{fig:align}
\end{figure}
Then, we saved the checkpoints for GraphCL~\cite{graphCL}, SimGRACE~\cite{simgrace}, and DC-GCL every 10 epochs during pre-training, as displayed in Fig.~\ref{fig:align}. We observed that all three methods are effective in reducing the alignment and uniformity metrics. However, as the pre-training progress, DC-GCL achieved lower alignment and uniformity than the two other baselines. Thus, compared to SimGRACE and GraphCL, which focus on a single perspective, the dual-perspective augmentation strategy effectively brings positive pairs closer in the embedding space. 

\section{Experiment}
\subsection{Unsupervised Learning}
\begin{table*}[!h]
\centering
\caption{Experimental results for \textbf{unsupervised representation learning} in graph classification task. The results for baseline methods are sourced from prior studies. -- indicates the absence of corresponding results in the original paper. \textbf{Bold} or \underline{underline} indicates the best or second-best result, respectively, among self-supervised methods.}
% \vspace{-0.5em}

\label{tab:un_res}
\renewcommand\arraystretch{1.3} % 行间距
\setlength\tabcolsep{6pt} % 列间距
\resizebox{0.99\textwidth}{!}{%
\begin{tabular}{@{}l|cccccccc@{}}
\toprule
\multirow{1}{*}{} & \multirow{1}{*}{\textbf{PROTEINS}} & \multirow{1}{*}{\textbf{D\&D}} & \multirow{1}{*}{\textbf{NCI1}} &  \multirow{1}{*}{\textbf{MUTAG}} & \multirow{1}{*}{\textbf{IMDB-B}} & \multirow{1}{*}{\textbf{COLLAB}} & \multirow{1}{*}{\textbf{RDT-B}}  & \multirow{1}{*}{\textbf{RDT-M5K}}   \\ \midrule
%\midrule
%\rowcolor{Gray}
GraphTrans~\cite{graphtrans} & $75.18_{\pm 3.36}$ & $75.24_{\pm 4.83}$ & $82.60_{\pm 1.20}$ & $87.22_{\pm 7.05}$ & $74.50_{\pm 2.89}$ & $79.81_{\pm 0.84}$ & $88.58_{\pm 1.30}$ & $56.06_{\pm 2.44}$\\
GraphGPS~\cite{graphgps}  & $75.77_{\pm 2.19}$ & $75.98_{\pm 1.53}$ & $84.21_{\pm 2.25}$  & $85.00_{\pm 3.16}$  & $77.40_{\pm 0.63}$  & $81.40_{\pm 0.26}$ & $88.40_{\pm 1.15}$ & $57.39_{\pm 1.04}$\\
\midrule
GL~\cite{gl} & -- & -- & -- &  $81.66_{\pm 2.11}$ & $65.87_{\pm 0.98}$ & -- & $77.34_{\pm 0.18}$ & $41.01_{\pm 0.17}$\\

WL~\cite{wl} & $72.92_{\pm 0.56}$ & -- & $80.01_{\pm 0.50}$ & $80.72_{\pm 3.00}$ & $72.30_{\pm 1.44}$ & -- & $68.82_{\pm 0.41}$ & $46.06_{\pm 0.21}$\\

DGK~\cite{dgk} & $73.30_{\pm 0.82}$ & -- & $\underline{80.31_{\pm 0.46}}$ & $87.44_{\pm 2.72}$ & $66.96_{\pm 0.56}$ & -- & $78.04_{\pm 0.39}$ & $41.27_{\pm 0.18}$\\
\midrule
%graph2vec~\cite{graph2vec}  & $73.30_{\pm 2.05}$ &  -- & $73.22_{\pm 1.81}$ & $83.15_{\pm 9.25}$ & $71.10_{\pm 0.54}$ &  --  & $75.78_{\pm 1.03}$ & $47.86_{\pm 0.26}$ \\

%MVGRL~\cite{mvgrl}               & --                                                 & --     & --  & $89.70_{\pm 1.10}$              & $74.20_{\pm 0.70}$                                            &--                       & $84.50_{\pm 0.60}$                    &--                      \\

Infograph~\cite{infoGraph}               & $74.44_{\pm 0.31}$                              &  $72.85_{\pm 1.78}$                    & $76.20_{\pm 1.06}$              & $89.01_{\pm 1.13}$                         & $73.03_{\pm 1.87}$                       &$70.65_{\pm 1.13}$                       & $82.50_{\pm 1.42}$       & $53.46_{\pm 1.03}$                          \\

GraphCL~\cite{graphCL}               & $74.39_{\pm 0.45}$                                 &   $\underline{78.62_{\pm 0.40}}$                    & $77.87_{\pm 0.41}$         &   $86.80_{\pm 1.34}$     & $71.14_{\pm 0.44}$                      & $71.36_{\pm 1.15}$                       &$89.53_{\pm 0.84}$                       & $55.99_{\pm 0.28}$                   \\

JOAO~\cite{joao}        & $74.55_{\pm 0.41}$                                          &    $77.32_{\pm 0.54}$      & $78.07_{\pm 0.47}$                             &  $87.35_{\pm 1.02}$              &  $70.21_{\pm 3.08}$                      &     $69.50_{\pm 0.36}$                 &  $85.29_{\pm 1.35}$          &   $55.74_{\pm 0.63}$                         \\ 
JOAOv2~\cite{joao}    & $74.07_{\pm 1.10}$                                          &    $77.40_{\pm 1.15}$      & $78.36_{\pm 0.53}$                             &  $87.67_{\pm 0.79}$              &  $70.83_{\pm 0.25}$                      &     $69.33_{\pm 0.34}$                 &  $86.42_{\pm 1.45}$          &   $56.03_{\pm 0.27}$      \\

%LP-Info~\cite{lp-info} & -- & -- & -- & -- & $71.90_{\pm 0.40}$ & $74.84_{\pm 0.31}$  & $87.81_{\pm 0.45}$ & $53.32_{\pm 0.33}$ \\
SimGRACE~\cite{simgrace}               & $75.35_{\pm 0.09}$                           &     ${77.44}_{\pm 1.11}$                       & $79.12_{\pm 0.44}$          &  $89.01_{\pm 1.31}$     &$71.30_{\pm 0.77}$                              &$71.72_{\pm 0.82}$                       & $89.51_{\pm 0.89}$     & $55.91_{\pm 0.34}$       \\
iGCL~\cite{igcl} & $74.8_{\pm 0.5}$ &   --  & $\textbf{82.7}_{\pm \textbf{0.4}}$  
& $\underline{89.8_{\pm 1.2}}$ & $\underline{72.6_{\pm 0.6}}$ & $72.0_{\pm 0.8}$ & -- &  --\\
DualGCL~\cite{dualgcl} & $\underline{75.6_{\pm 0.5}}$ & -- & ${79.8_{\pm 0.3}}$ & $88.5_{\pm 0.9}$ & --  & $\underline{75.3_{\pm 0.4}}$ & -- & -- \\
%CGKS~\cite{cgks} & $\underline{76.0_{\pm 0.2}}$ &   --  & $79.1_{\pm 0.2}$  & $\textbf{91.7}_{\pm \textbf{1.2}}$ & -- & $\underline{76.8_{\pm 0.1}}$ & $\textbf{91.7}_{\pm \textbf{0.2}}$ &  $\underline{57.0_{\pm 0.1}}$ \\
DRGCL~\cite{drgcl} & $75.2_{\pm 0.6}$ & $78.4_{\pm 0.7}$ & $78.7_{\pm 0.4}$ & $89.5_{\pm 0.6}$ & $72.0_{\pm 0.5}$ & $70.6_{\pm 0.8}$ & $\underline{90.8_{\pm 0.3}}$ & $\underline{56.3_{\pm 0.2}}$\\
\midrule
DC-GCL (ours)     & $\textbf{76.45}_{\pm \textbf{0.26}} $   &   $\textbf{79.43}_{\pm \textbf{0.63}}$ &  $79.62_{\pm 0.37}$       & $\textbf{90.50}_{\pm \textbf{1.55}}$      &                        $\textbf{73.16}_{\pm \textbf{0.45}}$ &   $ \textbf{{79.83}}_{\pm \textbf{0.43}}$   &  $\textbf{91.04}_{\pm \textbf{0.45}}$      & $\textbf{57.17}_{\pm \textbf{0.30}}$                      \\   \bottomrule
\end{tabular}
}
% \begin{minipage}{1.0\linewidth} \small % \footnotesize %\scriptsize %\scriptsize %\tiny
% \vspace{0.5em}
% Notations:  The results of baselines are from previous paper if avaliable.
% \end{minipage} 
% \vspace{-0.9em}
\end{table*}

\begin{table}[!h]
\centering
\renewcommand\arraystretch{1.25} % 行间距
\setlength\tabcolsep{5pt} % 列间距
\caption{Statistics of datasets of the \underline{unsupervised learning} task.}
\label{tab:un_dataset}
\resizebox{0.48\textwidth}{!}{%
\begin{tabular}{@{}c|lcrrc@{}}
\toprule
\textbf{Category}                  & \textbf{Datasets}  & \textbf{\# Graphs} & \textbf{\# Nodes} & \textbf{\# Edges} & \textbf{\# Classes} \\ \midrule
\multirow{2}{*}{Bioinfo.}
& PROTEINS & 1,113     & 39.06    & 72.82    & 2        \\ 
& D\&D & 1,178        & 284.32    & 715.66   & 2        \\
\midrule
\multirow{2}{*}{Molecules}                
& NCI1     & 4,110     & 29.87    & 32.30    & 2        \\
& MUTAG     & 188     & 17.9    & 19.79    & 2          \\
\midrule
\multirow{4}{*}{\begin{tabular}[c]{@{}c@{}}Social \\  Networks\end{tabular}} 
& IMDB-B   & 1,000     & 19.77    & 96.53    & 2 \\
& COLLAB   & 5,000     & 74.49    & 2,457.78  & 3      \\ 
& REDDIT-B & 2,000     & 429.63   & 497.75   & 2       \\
& REDDIT-M & 4,999     & 508.52   & 594.87   & 5       \\
\bottomrule
\end{tabular}%
}
\end{table}
\textbf{Dataset.} We utilized eight commonly-used real-world datasets from various sources to validate the proposed method's effectiveness. All the datasets were obtained the TU database~\cite{tu-dataset} (\textit{i.e.} NCI1, PROTEINS, MUTAG, D\&D, IMDB-BINARY, COLLAB, REDDIT-BINARY, and REDDIT-MULTI-5K), which includes diverse domains (\textit{e.g.} social networks, molecules, and bioinformatics), and the detailed dataset statistics are listed in Table~\ref{tab:un_dataset}. All the adopted datasets can be downloaded from PyTorch Geometric (PyG)~\cite{pytorch-geometric}.

\textbf{Baselines.} To demonstrate the proposed model's effectiveness during unsupervised learning, we compared DC-GCL with the following 12 baselines, including two GT-based supervised, three graph kernel, and seven contrastive methods: 
\begin{itemize}
    \item \textbf{The GT-based supervised methods}:\\
        \textbf{GraphTrans}~\cite{graphtrans} designed a novel GNNs readout module that uses a special token to aggregate all pairs interactions into a classification vector. It has been proven that modeling the node–node interaction pair is particularly important for large graph classification tasks. \\
        \textbf{GraphGPS}~\cite{graphgps} combines message passing networks with linear (i.e. long-range) transformer models to create a hybrid network.
    \item \textbf{The graph kernel methods}: \\
         \textbf{Graphlet kernel (GL)}~\cite{gl} compares graphs based on subgraph frequencies, offering insights into structural similarities. However, it can be computationally demanding for large graphs. \\
         \textbf{Weisfeiler-Lehman sub-tree kernel (WL)}~\cite{wl} uses subtree patterns for graph classification, extending the classical Weisfeiler-Lehman test to incorporate structural information beyond vertex labels. \\
         \textbf{Deep graph kernel (DGK)}~\cite{dgk} leverages the dependency information between sub-structures to define the similarities between graphs. Then, these similarities are used to define the latent representation of graph structures.
    \item \textbf{The contrastive methods}: \\
         \textbf{Infograph}~\cite{infoGraph} is a method for learning graph-level representations. It maximizes the mutual information between graph-level and substructure representations.\\
         \textbf{GraphCL}~\cite{graphCL} is a framework for learning
          unsupervised data representations using various graph augmentations. \\
         \textbf{JOAO}~\cite{joao} is a unified bi-level optimization framework that could automatically selects the augmentations, addressing the limitations of manual augmentation selection in GraphCL. \\
         \textbf{SimGRACE}~\cite{simgrace} is a novel framework GCL framework using the original and perturbed graphs as inputs for the two correlated encoders. \\
         \textbf{iGCL}~\cite{igcl} leverages augmentations in the latent space learned from a variational graph autoencoder to improve GCL method, generating augmentations that preserve semantics by reconstructing the graph's topological structure.\\
         \textbf{DualGCL}~\cite{dualgcl} effectively learns representations through an adaptive hierarchical aggregation process, a transformer-based aggregator, and a novel dual-channel contrastive system. \\
         \textbf{DRGCL}~\cite{drgcl} proposes a dimension principle-aware GCL method and introduces a learnable dimension principle acquisition network and a redundancy reduction constraint.\\
\end{itemize}
For each baseline, we utilized the recommended settings as official implementation in their own papers. 

\textbf{Implementation Details.} For unsupervised representation learning, we implemented DC-GCL using Python(3.11.0), Pytorch(2.1.0), scikit-learn(1.3.1), and Pytorch Geometric (2.4.0). Furthermore, all the experiments were conducted on a Linux server equipped with 8 NVIDIA A100s. We optimized the DC-GCL model using the Adam~\cite{adam} optimizer, with the following parameters: $\beta_1=0.9$, $\beta_2=0.999$, $\epsilon=1e^{-8}$. To minimize the introduction of excessive hyper-parameters, we fix the augmentation ratio as 0.2, hidden size as 64, pre-training epochs as 100. For other hyper-parameters selection, we search the encoder layer in the set\{1, 2, 3\}, learning rate in the set\{$10^{-3}, 5\times10^{-4}, 2.5\times10^{-4}$\}, batch size in the set\{8, 16, 32, 128\}. Furthermore, we assessed our proposed model's effectiveness by measuring its performance during graph classification tasks. First, we trained the DC-GCL model on the entire datasets to acquire graph representations. During the evaluation, we utilized a LIBSVM~\cite{svm} classifier, with hyper-parameters selected from the set \{$10^{-3}, 10^{-2}, ..., 1, 10$\}. We use 10-fold cross validation with five different seeds, reporting the average accuracy and variance across five random seeds as the evaluation metrics. 

\textbf{Experimental Results.} Table~\ref{tab:un_res} provides a comprehensive overview of DC-GCL's performance results. According to these results, the following observations can be made: \textbf{1) State-of-the-art-performance. }DC-GCL excelled with the eight datasets, outperforming the listed ten self-supervised and two GT-based supervised baselines. Specifically, DC-GCL achieved a notable accuracy improvement of 6.0\%, 1.1\%, and 1.5\% on the COLLAB, PROTEINS, and RDT-M5K datasets, respectively. Additionally, DC-GCL's competitive performance underscores the importance of integrating a dual-perspective augmentation and employing multiple contrastive loss functions during pre-training. \textbf{2) Comparison with contrastive methods. }DC-GCL outranked all seven traditional contrastive learning methods across the eight benchmark datasets. This validates the significance of the dual-perspective augmentation and multiple contrastive loss modules. \textbf{3) Comparison with graph kernel methods:} Through comparison with three traditional graph kernel methods as the baselines, we observed that DC-GCL demonstrates significant improvements across most datasets, except for the NCI1 dataset. For example, on the RDT-B dataset, DC-GCL achieved an improvement of 16.7\%. \textbf{4) Comparison with supervised methods: }To further evaluate the DC-GCL architecture's effectiveness, we selected two GT-based models as baselines for comparison. On the benchmark PROTEINS, DD, and MUTAG dataset, DC-GCL achieved improvements of 0.9\%, 4.5\%, and 3.8\%, respectively, showing the potential of GT-based self-supervised learning methods for graph representation learning. \textbf{5) Superior performance on dense graphs: }DC-GCL exhibited significant improvements on denser graphs, such as D\&D, COLLAB, RDT-B, and RDT-M, consistently securing a top-two ranking among all baselines. Notably, on the COLLAB and RDT-M5K datasets, DC-GCL demonstrated a notable improvement of 6.0\% and 1.5\% respectively. This indicates that DC-GCL's capability to learn dense graph representations and effectively capture their structural features is outstanding, leading to higher performance. 
\subsection{Transfer Learning}
\textbf{Dataset.} To assess the proposed approach's transferability, we evaluated its transfer learning performance for molecular property prediction, adhering to the experimental setups from previous studies~\cite{pretraingnn,graphCL,simgrace}. We pre-trained the model with 2 million unlabeled molecules sampled from the ZINC15 dataset~\cite{zinc15}. For the fine-tuning phase, we adopted eight classification datasets from the MoleculeNet database~\cite{moleculenet}. The specific datasets are summarized in the Table~\ref{tab:trans_dataset}. 
\begin{table*}[!t]
\centering
\caption{Experimental results for \textbf{transfer learning} on molecular property prediction task. The model is initially pre-trained on the ZINC15 dataset and subsequently fine-tuned on the above datasets. The reported metrics are ROC-AUC scores. \textbf{Avg.} denotes the average performance. The results for baseline methods are derived from previous studies. $^*$ indicates the results borrowed from DRGCL.}
\label{tab:transfer_res}
\renewcommand\arraystretch{1.3} % 行间距
\setlength\tabcolsep{4pt} % 列间距
\resizebox{0.9\textwidth}{!}{%
\begin{tabular}{@{}l|cccccccc|c@{}}
\toprule
            & \textbf{BBBP}                                  & \textbf{Tox21}                                  & \textbf{ToxCast}                               & \textbf{SIDER}                                 & \textbf{ClinTox}                                  & \textbf{MUV}                          & \textbf{HIV}                                   & \textbf{BACE}                                                       & \textbf{Avg.}               \\ \midrule
No-pretrain & $69.1{\pm 2.4}$ & $74.6{\pm 0.8}$ & $63.1{\pm 0.7}$ & $58.1{\pm 1.0}$ & $64.4{\pm 4.2}$ & $71.2{\pm 2.0}$ & $76.0{\pm 1.4}$ & $77.0{\pm 2.1}$ & 69.2 \\
\midrule
ContextPred~\cite{pretraingnn} & $64.3 {\pm 2.8}$                        & $75.7 {\pm 0.7}$             & $63.9 {\pm 0.6}$                        & $60.9 {\pm 0.6}$                        & $65.9 {\pm 3.8}$                           & $75.8 {\pm 1.7}$               & $77.3 {\pm 1.0}$                        & \multicolumn{1}{c|}{$\underline{79.6 {\pm 1.2}}$}                        & 70.4               \\
AttrMasking~\cite{pretraingnn} & $64.3 {\pm 2.8}$                        & ${76.7} {\pm 0.4}$  & $64.2 {\pm 0.5}$ & $61.0 {\pm 0.7}$            & $71.8 {\pm 4.1}$                           & $74.7 {\pm 1.4}$               & $77.2 {\pm 1.1}$                        & \multicolumn{1}{c|}{$79.3 {\pm 1.6}$}                        & 71.1               \\
Infomax~\cite{pretraingnn}     & $68.8 {\pm 0.8}$                        & $75.3 {\pm 0.5}$                         & $62.7 {\pm 0.4}$                        & $58.4 {\pm 0.8}$                        & $69.9 {\pm 3.0}$                           & $75.3 {\pm 2.5}$               & $76.0 {\pm 0.7}$                        & \multicolumn{1}{c|}{$75.9 {\pm 1.6}$}                        & 70.3               \\
GraphCL~\cite{graphCL} & $69.7 {\pm 0.7}$                        & $73.9 {\pm 0.7}$                         & $62.4 {\pm 0.6}$                        & $60.5 {\pm 0.9}$                        & $76.0 {\pm 2.7}$                           & $69.8 {\pm 2.7}$               & $\underline{78.5 {\pm 1.2}}$ & \multicolumn{1}{c|}{$75.4{ \pm 1.4}$}                        & 70.8               \\
JOAO~\cite{joao}        & $70.2 {\pm 1.0}$                        & $75.0 {\pm 0.3}$                         & $62.9 {\pm 0.5}$                        & $60.0 {\pm 0.8}$                        & $\underline{81.3 {\pm 2.5}}$               & $71.7 {\pm 1.4}$               & $76.7 {\pm 1.2}$                        & \multicolumn{1}{c|}{$77.3 {\pm 0.5}$}                        & 71.9               \\
ADGCL$^*$~\cite{adgcl}  & $70.5 {\pm 1.8}$ &  ${74.5 {\pm 0.7}}$ &  $63.0 {\pm 0.5}$ &  $59.1 {\pm 0.9}$ &  $78.5 {\pm 3.7}$ &  $71.5 {\pm 2.2}$ &  $75.9 {\pm 1.4}$ &  $74.0 {\pm 2.2}$ & 70.9\\
SimGRACE~\cite{simgrace} & $\underline{71.3 {\pm 0.9}}$ &  $\underline{75.6 {\pm 0.5}}$ &  $63.4 {\pm 0.5}$ &  $60.6 {\pm 1.0}$ &  $75.6 {\pm 3.0}$ &  $\textbf{76.9} {\pm \textbf{1.3}}$ &  $75.2 {\pm 0.9}$ &  $75.0 {\pm 1.7}$ & 71.7\\
RGCL$^*$~\cite{rgcl}  & $\textbf{71.4} {\pm \textbf{0.9}}$ & $75.2 {\pm 0.3}$ & $63.3 {\pm 0.2}$  & $\textbf{61.4} {\pm \textbf{0.6}}$ & $76.4 {\pm 3.4}$             & $72.6 {\pm 1.5}$   & $77.9 {\pm 0.8}$            & ${76.0} {\pm {0.8}}$ & $71.8$ \\
%LP-info~\cite{lp-info} & $\textbf{71.4} {\pm \textbf{0.5}}$ & $74.5 {\pm 0.5}$ & $63.0 {\pm 0.3}$    & $59.7 {\pm 0.4}$ & $74.8 {\pm 2.1}$             & $73.0 {\pm 2.3}$   & $77.0 {\pm 1.1}$            & $\textbf{80.2} {\pm \textbf{1.4}}$ & $71.7$ \\

DRGCL~\cite{drgcl} & $71.2 {\pm 0.5}$ & $74.7 {\pm 0.5}$ & $\underline{64.0 {\pm 0.5}}$ &$\underline{61.1 {\pm 0.8}}$ & $78.2 {\pm 1.5}$ & $73.8 {\pm 1.1}$ & $\textbf{78.6} {\pm \textbf{1.0}}$ & $78.2 {\pm 1.0}$ & $\underline{72.5}$\\
\midrule 
DC-GCL (ours)   & $70.6 {\pm 0.8}$           & $\textbf{75.8} {\pm \textbf{0.6}}$             & $\textbf{65.3} {\pm \textbf{0.5}}$            & ${60.6 {\pm 0.6}}$                        & $\textbf{82.7}{\pm \textbf{3.6}}$ & $\underline{76.0 {\pm 1.5}}$ & $77.8 {\pm 0.8}$                        & $\textbf{79.9} {\pm \textbf{1.7}}$                                  & $\textbf{73.5}$ 
\\
\bottomrule
\end{tabular}%
}
\end{table*}

\begin{table}[!h]
\centering
\renewcommand\arraystretch{1.15} % 行间距
\caption{Statistics of datasets of the \underline{transfer learning} task.}
\label{tab:trans_dataset}
\resizebox{0.48\textwidth}{!}{%
\begin{tabular}{@{}l|crrrc@{}}
\toprule
\textbf{Dataset} &
  \textbf{Utilization} &
  \textbf{\# Graph} &
  \textbf{\# Nodes} &
  \textbf{\# Edges} &
  \textbf{\# Classes} \\ \midrule

ZINC & Pre-training & 2,000,000 & 26.62 & 57.72 & -- \\
\midrule
BBBP & Fine-tuning  & 2,039 & 24.06 & 51.90  & 1\\
Tox21 & Fine-tuning   & 7,831 & 18.57 & 38.59 & 12\\ 
ToxCast  & Fine-tuning & 8,576 & 18.78 & 38.52 & 617 \\
SIDER  & Fine-tuning & 1,427 & 33.64 & 70.71  & 27\\
ClinTox  & Fine-tuning & 1,477 & 26.16 & 55.77 & 2 \\
MUV & Fine-tuning & 93,087 & 24.23 & 52.56 & 17 \\ 
HIV & Fine-tuning & 41,127 & 22.5 & 27.5 & 1 \\ 
BACE & Fine-tuning & 1,513 & 34.09 & 73.72 & 1 \\ 
\bottomrule
\end{tabular}%
}
\end{table}

\textbf{Baselines.} We evaluated transferability effectiveness of DC-GCL by comparing it with baselines, including Infomax, AttrMasking and ContextPred~\cite{pretrain-gnn}, and state-of-the-art contrastive methods, such as GraphCL~\cite{graphCL}, JOAO~\cite{joao}, SimGRACE~\cite{simgrace}, AD-GCL~\cite{adgcl}, and RGCL~\cite{rgcl}, and DRGCL~\cite{drgcl}. The baselines not introduced in the above text will be detailed in the following text:
\begin{itemize}
    \item \textbf{Infomax}~\cite{pretrain-gnn} trains a node encoder that maximizes mutual information between local node representations and a pooled global graph representation.
    \item \textbf{AttrMasking}~\cite{pretrain-gnn} captures domain knowledge by learning the regularities of the node/edge attributes distributed over the graph structure.
    \item \textbf{ContextPred}~\cite{pretrain-gnn} uses subgraphs to predict their surrounding graph structures, aiming to pre-train an encoder so that it maps nodes appearing in similar structural contexts to the nearby embedding.
    \item \textbf{AD-GCL}~\cite{adgcl} enables GNNs to avoid capturing redundant information during the training by optimizing adversarial graph augmentation strategies used in GCL.
    \item \textbf{RGCL}~\cite{rgcl} uses a rationale generator to reveal salient features about the graph, with instance-discrimination as the rationale. Then, it creates rationale-aware views for contrastive learning.
\end{itemize}
For each baseline, we utilized the experimental results as reported in their original papers. For the cases where results were unavailable, we conducted the experiment using settings identical to those of our proposed model.

\textbf{Implementation Details.} For transfer learning, we implement DC-GCL with Python (3.9), Pytorch (1.12.0), scikit-learn (1.3.1), rdkit (2022.03.2), and Pytorch Geometric (2.2.0). For the pre-training phase, we set the augmentation ratio as 0.2, batch size as 256, hidden size as 300, learning rate as 0.001. We searched the encoder layer in the set \{1, 2, 3\}. For the fine-tuning phase, we fixed the coarsening layer the same as pre-training. In addition, we searched for the learning rate in the set \{0.002, 0.001, 0.0005\}, and batch size in the set \{32, 64, 128\}. Furthermore, we used the PE masking strategy for data augmentation and drop head strategy for model augmentation in transfer learning. We pre-train the model 100 epochs and saved the model every 20 epochs. For the evaluation, we partitioned the downstream datasets into training, validation, and testing sets with an 80/10/10\% split using scaffold-split. We employed ten different random seeds and report average ROC-AUC scores and their standard deviations. 

\textbf{Experimental Results.} Table~\ref{tab:transfer_res} presents the comprehensive results of the transfer learning experiments. Based on the results, we can observe: \textbf{1) State-of-the-art performance. }DC-GCL achieved a 1.4\% improvement in average performance metrics, with five out of eight datasets ranking within the top two. This consistent high performance across the majority of datasets underscores the effectiveness of DC-GCL in molecule datasets. \textbf{2) Superior of the DC-GCL Architecture. }Compared with the single augmentation methods such as GraphCL and SimGRACE, DC-GCL exhibited significant improvements of 3.7\% and 2.4\%, for the average performance metrics, respectively. This suggests that the dual-perspective augmentation module can enhance the model's performance by generating more positive sample pairs and controllable novel augmentation strategies can generate more reliable positive samples, aligning with previous unsupervised representation learning experiments.
\begin{figure*}[!t] % !htb
\begin{center}
\includegraphics[width=1.0\linewidth]{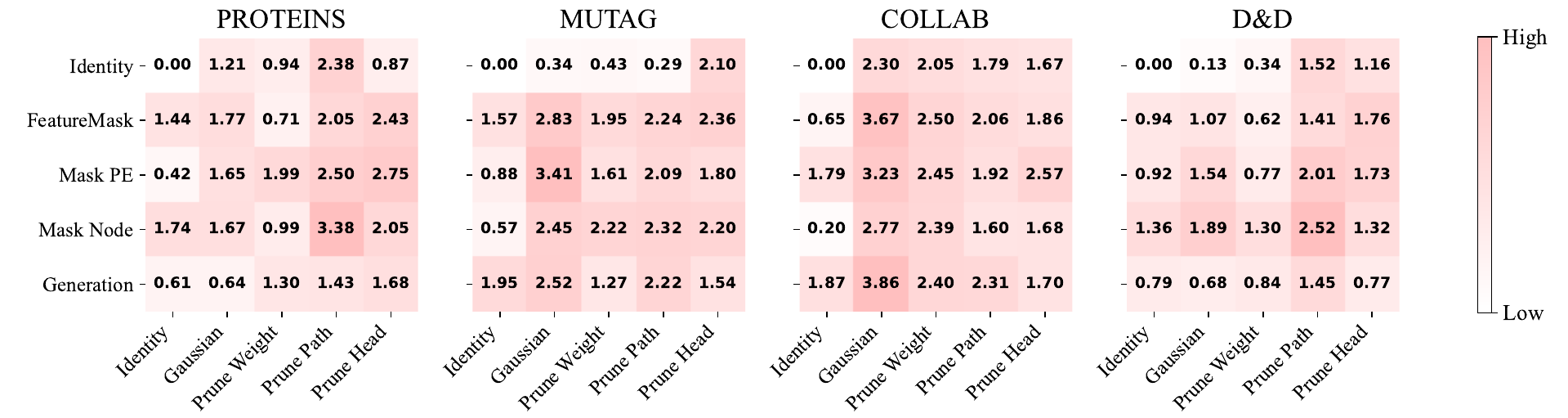}
\end{center}
\caption{\textbf{Analysis of the Augmentation Strategy.} We report unsupervised learning accuracy improvement(\%) when contrasting different combination of data (vertical axis) and model (horizontal axis) augmentation. Deeper colors indicate better performance improves. ``Identity'' denotes without any augmentation methods for contrastive learning. 
}
\label{fig:augmentation}
\end{figure*}

% dual分析一下不管是从data和model都一致的比dual低。
\subsection{Ablation Study}
\textbf{Ablation on Components of DC-GCL.} We conducted a thorough analysis to assess the impact of various components within the DC-GCL framework. We controlled for each experiment by altering only one variable at a time, ensuring a systematic examination of the specific component's impact. The experimental results presented at Table~\ref{tab:ablation} offer valuable insights into the significance of each component: \textbf{1) Dual-perspective augmentation:} In both data and model augmentation, the effectiveness of single-perspective augmentation consistently performs poorly when compared to dual-perspective augmentation. The positive results underscore the efficacy of applying dual-perspective augmentation to GCL architecture. This augmentation strategy significantly enhances positive sample diversity and provides more information, thereby improving the model's ability to capture invariant representations. \textbf{2) Multi-view contrastive loss: }The noticeable declines in performance observed in the absence of the multi-view contrastive component emphasized its pivotal role in enhancing the model's efficacy. The model may have difficulty learning consistency when the positive pair features are too far apart within the embedding space such as pairs ($\boldsymbol{z}^1$, $\boldsymbol{z}^4$), which were generated from different data or model augmentation. Specifically, on the PROTEINS dataset, we observed a reduction of 2.7\%, 1.0\% on D\&D, and 3.5\% for COLLAB. \textbf{3) GT-based encoder:} Replacing the GNN encoder with GTs led to a remarkable enhancement in performance, a phenomenon that becomes especially evident across the three aforementioned datasets. These results reaffirm the advantages of GTs over GNNs in contrastive learning and highlight their potential to significantly enhance the contrastive learning model performance.

\begin{table}[!t]
\centering
\caption{Ablation study of DC-GCL components. ``w/o Multi-View Contrastive'' means that we calculate the contrastive loss in arbitrary two views.}
\label{tab:ablation}
\renewcommand\arraystretch{1.25} % 行间距
\setlength\tabcolsep{3pt} % 列间距
\resizebox{0.49\textwidth}{!}{%
\begin{tabular}{l|ccc}
\hline
                  & PROTEINS          & D\&D              & COLLAB \\ \hline
\textbf{DC-GCL }          & $\textbf{76.45}_ {\pm \textbf{0.26}}$  & $\textbf{79.43}_ {\pm \textbf{0.63}}$  & $\textbf{79.83}_ {\pm \textbf{0.43}}$  \\ \hline
\quad w/o Data Augmentation & $\underline{75.59_ {\pm 0.49}}$  & $\underline{78.92_ {\pm 0.40}}$  & $\underline{78.74_ {\pm 0.30}}$  \\
\quad w/o Model Augmentation      & $74.99_ {\pm  0.49}$ & $78.47_ {\pm 0.56}$  & $78.33_ {\pm 0.46}$  \\
\quad w/o Multi-View Contrastive       & $74.43_ {\pm  0.39}$ & $78.63_ {\pm  0.33}$ & $77.12_ {\pm  0.47}$ \\
 \quad w/o GT-based Encoder   & $74.16_ {\pm  0.53}$ & $77.68_ {\pm  0.47}$ & $77.66_ {\pm  0.22}$ \\ 
\hline
\end{tabular}%
}
\end{table}

\textbf{Ablation on Augmentation Strategy.} To further study the effectiveness of the strategies we applied to contrastive learning, we used FeatureMask for data augmentation and Gaussian noise for model augmentation from previous works for comparison and detailed analysis. The results are displayed in Fig.~\ref{fig:augmentation}. We observed: \textbf{1) The superior of the dual-perspective augmentation module. }Consistent with previous experimental results, the dual-perspective augmentation strategy demonstrates a noticeable improvement in contrastive learning compared to the single augmentation strategy. In contrast, without augmentation, GCL's performance declines significantly. This observation aligns with our assumptions. Furthermore, if the positive pairs stay within the same position in the embedding space, the model will be unable to distinguish them, restricting its generalization for the unseen data. \textbf{2) The effectiveness of novel augmentation strategies in DC-GCL. }We introduced three controllable data and pruning-based model DC-GCL augmentation strategies, showcasing significant improvement over the traditional approaches in previous works. For example, for the MUTAG dataset, new proposed strategies (\textit{i.e.} drop node+drop path) acheived an improvement of 1.6\% than the previous strategies (\textit{i.e.} feature mask+Gaussian noise).

\subsection{Parameter Analysis}
\begin{figure}[!t] % !htb
\begin{center}
\includegraphics[width=1.0\linewidth]{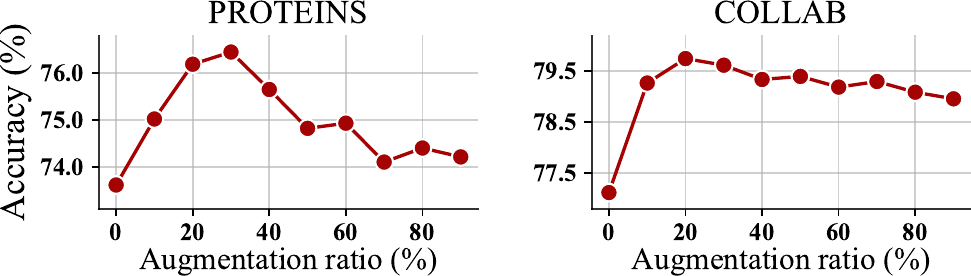}
\end{center}
\caption{\textbf{Analysis of the Augmentation Ratio(\%).} Performance of various augmentation ratios. We choose the best augmentation strategy each dataset for analysis. 
}
\label{fig:ratio}
\end{figure}
We conducted a detailed analysis to explore the effect of augmentation ratio variation on DC-GCL's performance. The results are illustrated in the Fig.~\ref{fig:ratio}. Upon analyzing the results, the optimal augmentation ratio varies depending on the dataset. We observed that as the augmentation ratio increases, the performance level improves. This indicates that the positive pairs are too close within the embedding space, making it difficult for the model to distinguish between them when the augmentation ratio is relatively low. However, when the ratio is beyond the threshold, accuracy gradually declines, suggesting that excessively high augmentation ratio can result within positive samples being too widely dispersed in the embedding space. This compromises the semantic information, hindering the model's ability to learn representations. To minimize the introduction of excessive hyper-parameters, we set the ratio as 0.2 throughout the experiment.

\subsection{Efficiency Analysis}
We analyzed DC-GCL's complexity. As we mentioned above, DC-GCL adopts GT as an encoder instead of the GNNs used in previous GCL methods. In contrast to the neighbor aggregation operation in GNNs, the self-attention mechanism in GTs significantly reduces the model's computational efficiency. Thus, the time complexity from $\mathcal{O}(|E|)$ increases to $\mathcal{O}(n^2)$. Additionally, we used the dual-perspective augmentation strategy to generate four correlated graphs, and inserted them into the encoder, thus, the final time complexity is approximately $\mathcal{O}(4n^2)$. The detailed time cost statistics are presented in Table~\ref{tab:time_cost}. Upon analysis, we observed that the DC-GCL model, which employs GTs as an encoder and adopts dual-perspective augmentation, exhibited a higher time consumption than the GNN-based methods and single-perspective augmentation such as GraphCL and SimGRACE. However, when compared with other contrastive learning methods, such as DRGCL~\cite{drgcl}, the DC-GCL's time consumption was consistently lower, indicating that the its time cost is controllable. 
\begin{table}[!t]
\centering
\renewcommand\arraystretch{1.25} % 行间距
\setlength\tabcolsep{5pt} % 列间距
\caption{Time consumption (Seconds) of 100 epochs for other three GNN-based baselines and our proposed DC-GCL across four datasets.}
\label{tab:time_cost}
\resizebox{0.45\textwidth}{!}{%
\begin{tabular}{@{}l|rrrr@{}}
\toprule
 & \textbf{MUTAG} & \textbf{NCI1} & \textbf{IMDB-B} & \textbf{PROTEINS}\\ \midrule
GraphCL~\cite{graphCL} &5.6s & 84.0s & 15.4s & 28.2s \\
SimGRACE~\cite{simgrace} &6.7s & 96.4s & 17.3s & 33.4s \\
DRGCL~\cite{drgcl} & 175.4s & 5243.1s & 581.2s & 731.2s \\
\midrule
DC-GCL & 18.7s & 196.4s & 50.2s & 223.7s \\
\bottomrule
\end{tabular}%
}

\end{table}

\section{Conclusion}
In this study, we introduced the DC-GCL architecture, including dual-perspective augmentation and multi-view contrastive loss modules. We also proposed three strategies to generate more reliable positive samples for the data and model augmentation perspectives. Through comprehensive experiments covering two distinct graph learning tasks, DC-GCL demonstrated superior performance compared to the single augmentation strategy. Furthermore, DC-GCL consistently demonstrated its ability to distinguish between various positive sample pairs and exhibited superior generalization capabilities. Then, we conducted ablation studies to demonstrate our proposed module's effectiveness, which can be transferred to other contrastive learning methods. Despite its competitive performance, DC-GCL still has various areas for further improvement, including \textbf{1)} exploring feasibility of integrating the contrastive and generative learning, and \textbf{2)} investigating if negative samples can be generated when the interference exceeds a certain threshold.

\bibliographystyle{IEEEtran}
\bibliography{TKDE/tkde}

\end{document}